\documentclass[conference,a4paper]{IEEEtran}
\usepackage[utf8]{inputenc}
\usepackage{cite}
\usepackage{amsmath}

\usepackage{graphicx}
\graphicspath{ {./img/} }

\usepackage{authblk}
\usepackage{todonotes}
\title{Color Cerberus}

\author[1]{A.~Savchik}
\author[1]{E.~Ershov}
\author[1,2]{S.~Karpenko}
\affil[1]{Institute for Information Transmission Problems of the Russian Academy of Sciences (IITP RAS)}
\affil[2]{Glidewell.io}
\date{}                     %% if you don't need date to appear
\begin{document}

\maketitle

\begin{abstract}
    Simple convolutional neural network was able to win ISISPA color constancy competition. Partial reimplementation of \cite{bianco2017single} neural architecture would have shown even better results in this setup.
\end{abstract}

\section{Introduction}

During the last decade, using convolutional neural network as a baseline for almost any problem in computer vision has became an industry standard. Time after time and setup after setup even relatively small convolutional networks learned on medium size datasets demonstrate excellent performance in a wide variety of tasks.

The goal of ISISPA color constancy competition was to predict the one "main" light source spectrum from an image. To be more precise, simple three dimensional lighting model was used. About 2000 images were given, each of them having grey cube posed into the scene. All images were taken with the same camera (but in different modes, see below). The task was to find three numbers, corresponding to pixel colors of some side of the grey cube. Of course, during inference a zone with the cube has to be masked.

Following the aforementioned approach, we have trained classic LeNet architecture for this regression task. The winning architecture has several tweaks as described in the last section of the article.

As far as we know, CNN for color constancy was first successfully tested in Bianco et al in \cite{bianco2015color} where it was applied to reprocessed Gehler dataset \cite{colorchecker}. Interestingly enough, that network consisted of only one convolutional layer followed by a single hidden fully connected layer.

That work was superseded by \cite{bianco2017single}, with a slightly deeper net as a local illuminant estimator. It also has a sophisticated averaging procedure. After our LeNet-like net have been submitted but before the ground truth have become available we performed several experiments with local estimator from Bianco et al. architecture. Now we know that it would have shown even better results in the ISISPA contest. 

We also experimented with modified  Bianco et al. architecture and achieved even better results - however this improvement is not statistically significant and requires further investigations.

We start with analysis of datasets available in the field of color constancy. Then we report our learning procedure and the winning network architecture. We conclude with few general remarks on CNN applicability to color constancy tasks.
\section{Datasets}

\subsection{Collecting}

During the contest we've tried to leverage some publicly available datasets in addition to Cube+ \cite{banic2017unsupervised}. 
such as INTEL-TUT \cite{aytekin2017data}, NUS \cite{cheng2014illuminant}, Grey Ball \cite{ciurea2003large} and Color Checker \cite{hemrit2018rehabilitating}. 

At first glance one of the most promising datasets was INTEL-TUT.
First of all its huge (about 100GB), moreover authors declare that its purpose is to be the base for camera-independent color constancy solution development.
In fact it has just 3 different cameras: 2 photo cameras and 1 smartphone camera. 
Interesting fact, that INTEL-TUT images doesn't contain any color calibrating object in the scene.
On one hand it helps to eliminate problems with calibration object and tripod masking, on the other hand we loose opportunity to re-check markup consistency.
The size of dataset is huge just because its contains huge RAW files (50-100 MB each), the number of different images is approximately 1000.
Despite the fact that dataset has corresponding paper it still remains poorly documented: it is hard to understand how authors measure ground truth for non-laboratory cases.

\begin{figure}
    \centering
    \includegraphics[scale=0.35]{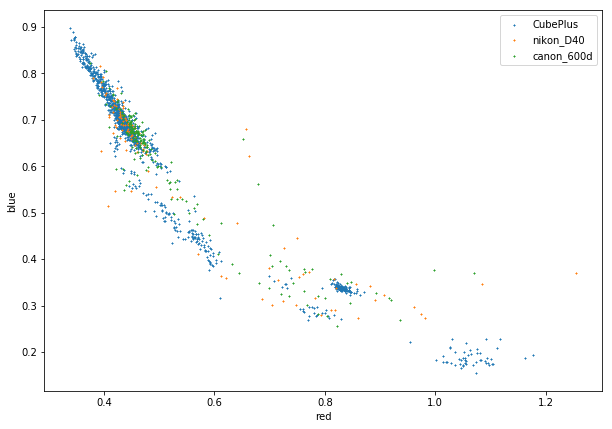}
    \caption{Chromaticities distribution on two-dimensional plane, where horizontal axis is R/G, while vertical B/G.}
    \label{fig:similar_sources}
\end{figure}

The second promising dataset was NUS, because it contains images captured by the same sensor (Canon 550D and 600D have the same matrix and default lens).
Indeed, we found two really close subdatasets to Cube+ that is ones collected with Canon 600D and Nikon D40, as shown in figure \ref{fig:similar_sources}.

\begin{figure}
    \centering
    \includegraphics[scale=0.35]{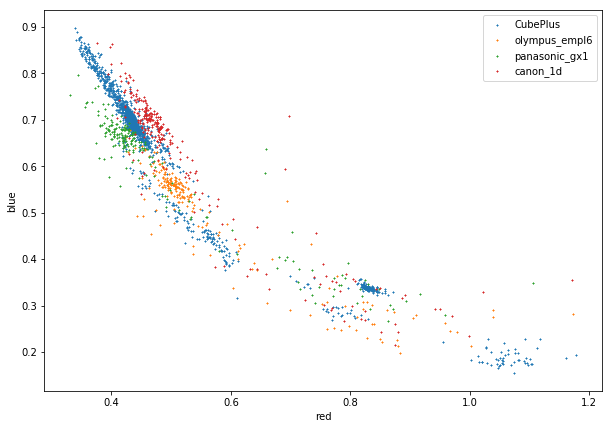}
    \caption{Chromaticities distribution on two-dimensional plane, where horizontal axis is R/G, while vertical B/G.}
    \label{fig:not_so_far_sources}
\end{figure}

Also we've selected a bunch of subdatasets which is a bit less similar with Cube+. 
These was images captured by Olympus EMLI6, Panasonic GX1, Canon 1D, see figure \ref{fig:not_so_far_sources}.

\begin{figure}
    \centering
    \includegraphics[scale=0.35]{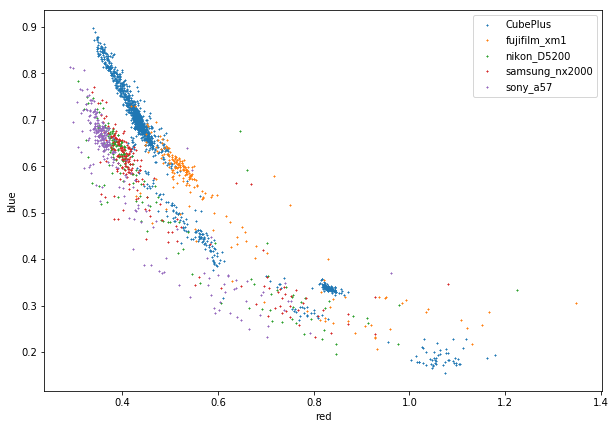}
    \caption{Chromaticities distribution on two-dimensional plane, where horizontal axis is R/G, while vertical B/G.}
    \label{fig:hardly_differ_sources}
\end{figure}

The rest cameras highly differ from Cube+ dataset, see figure \ref{fig:hardly_differ_sources}.
Note that chromaticities distribution for each camera has pretty similar structure: it is possible to select outdoor and indoor clusters, so maybe it is possible to compute some good projective transform between two distribution on chromaticity plane to augment the data, but we haven't try this idea.

The rest two datasets (ColorChecker, GreyBall) was downloaded, but we haven't use them during context, nevertheless these two datasets gives really good opportunity to estimate our solution performance and in some sense Cube+ representativity.

\subsection{Cube+ augmentation}

The other way to obtain more data for training is the color augmentation: we always can multiply each image pixel and estimated source vector element-wise with 3D vector, which in RGB (or ``zonal'') model equivalent to source color transformation.

So, theoretically we can get as many image from relatively small dataset as we want. 
But, there is some peculiarity with saturation: it is normal to multiply saturated pixel by value greater than 1 (they are already saturated), but reverse operation isn't correct.
If we divide saturated pixel with some value we will get unnatural image behaviour: saturated edge on histogram will be located far from image maximum.

\begin{figure}[ht]
    \centering
    \includegraphics[scale=0.6]{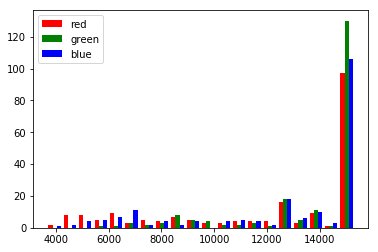}
    \caption{Histogram of image maximums for each channel.}
    \label{fig:max_hist}
\end{figure}

It leads us to problem: how to evaluate saturation value for particular image?
Authors of the dataset write on the corresponding web page: ''To make a conclusion about the maximum allowed intensity values of non-clipped pixels in the dataset images, histograms of intensities for various images were observed. 
If m is the maximum intensity for a given dataset image in any of its channels, then the best practice is to discard all image pixels that have a channel intensity that is greater than or equal to $m-2$''.
But if we look at the histogram of maximum of images for each channel, we will see  interesting behaviour: there are some peaks near $12600$, $13600$ and $15300$, see figure \ref{fig:max_hist}.

Indeed, maximum value seems to vary from image to image.
We've parsed EXIF files, provided by organizers and figure out that images was captured with different ISO level. 
That means that sensitivity setup differ from image to image and as it turned out it affect on the saturation level.
Finally we've figured out that images ISO and saturation level is hardly connected as shown in table \ref{tab:sat_vals}.
So, using this meta data we can overcome problem with saturation level.

\begin{table}[ht]
\centering
\begin{tabular}{|l|l|}
    \hline
    Saturation value & ISO \\
    \hline
    12652 & 160, 320, 640, 1250 \\
    \hline
    13584 & 100, 125 \\
    \hline
    15306 & 200, 250, 400, 500, 800, 1000, 1600 \\
    \hline
    15324 & 6400 \\ \cline{1-2}
\end{tabular}
\caption{Correspondence between saturation value and ISO level for Canon 550D.}
\label{tab:sat_vals}
\end{table}

Unfortunately, such augmentation didn't help. 
Probably this is because we haven't taken into account the type of a scene during augmentation, which could corrupt some semantic content of the dataset.
For example, outdoor scene probably wont be illuminated by greenish light.
So, may be it is reasonable to restrict the difference between real source chromaticity and generated or somehow take into account scene type.

\subsection{Dataset and contest metric features}

The main metric of success in this competition was median of reproduction error \cite{finlayson2014reproduction}:
\begin{equation}
    E(t, p) = \arccos{ \Bigg( \frac{ \langle p / t, [1,1,1] \rangle }{|p / t| \cdot \sqrt{3}} \Bigg)},
\end{equation}
where $t$, $p$ -- ground truth and estimated source chromaticity correspondingly, sign ``$/$'' means element-wise division, and ``$\langle \cdot, \cdot \rangle$'' means dot product.

The median in the score allows us to make huge mistakes for half of all images in test dataset without any penalty.
For example we can unfettered ignore all night and indoor images and work only with outdoor dataset (outdoor subset is bigger than union of the rest images).
Finally we've found that the constant value $(0.17, 0.40, 0.27)$ has score $2,03$, which is better than gray world score.
This value is located in the middle of outdoor cluster located in the upper left corner in scatter plot \ref{fig:similar_sources}.

Despite the rest datasets, Cube+ uses SpyderCube object to perform illuminance chromaticity evaluation.
The shape and color of the SpyderCube allows to retrieve two different estimation of the illumination (two white cube edges) chromaticities in the scene.
In fact, both estimation is a mixture of all light sources in the scene, that is why organizers provide ground truth for both edges of the cube, moreover they've performed selection of the dominant edge, i.e. dominant light source.
In our solution we've made the attempt to utilize this information.
\section{Solution}

During the contest we've performed approximately $150$ training experiments, tested a few different architectures, including several convolutional neural networks and U-Net-like networks with global answer averaging.
Also we've experimented with data augmentation, online patch selection, two downscale methods, different patch sizes and $L_2$ regularization coefficients.

Training dataset was formed by randomly selected $1390$ images from the Cube+ dataset.
For all training images we've set ROI as $x < 1900$ to exclude SpyderCube.
Then we've picked $100$ random patches on each image (every patch was a square with random size from the list: $64 \times 12$, $64 \times 18$, $64 \times 24$), after that each patch was downscaled with downlscale\_local\_mean function \cite{scikit_local_mean} to the shape $64 \times 64 \times 3$. 
The rest $317$ images from Cube+ (totally it contains $1707$ images) were used to form validation datasets (we've picked and downscaled a single random $64 \times 24$ patch from each image): val1 and val2.

\subsection{What have we submit?}

\begin{figure}
    \centering
    \includegraphics[scale=0.25]{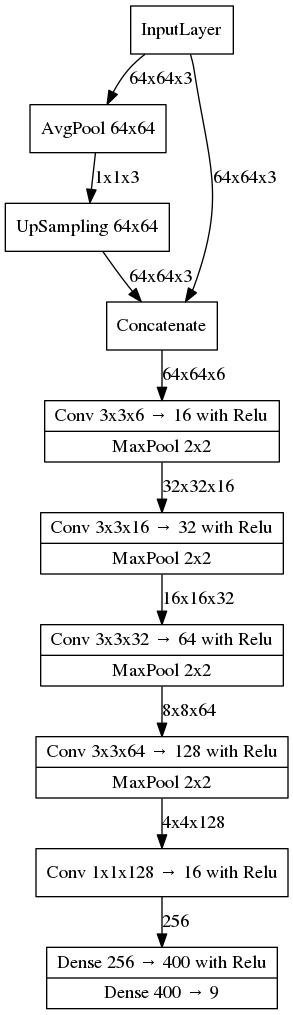}
    
    \caption{Architecture of the winning network. AvgPool+UpSampling branch calculates the GrayWorld method and being concatenated with the input image.}
    \label{fig:win_arch}
\end{figure}

\begin{figure}
    \centering
    \includegraphics[width=0.5\textwidth]{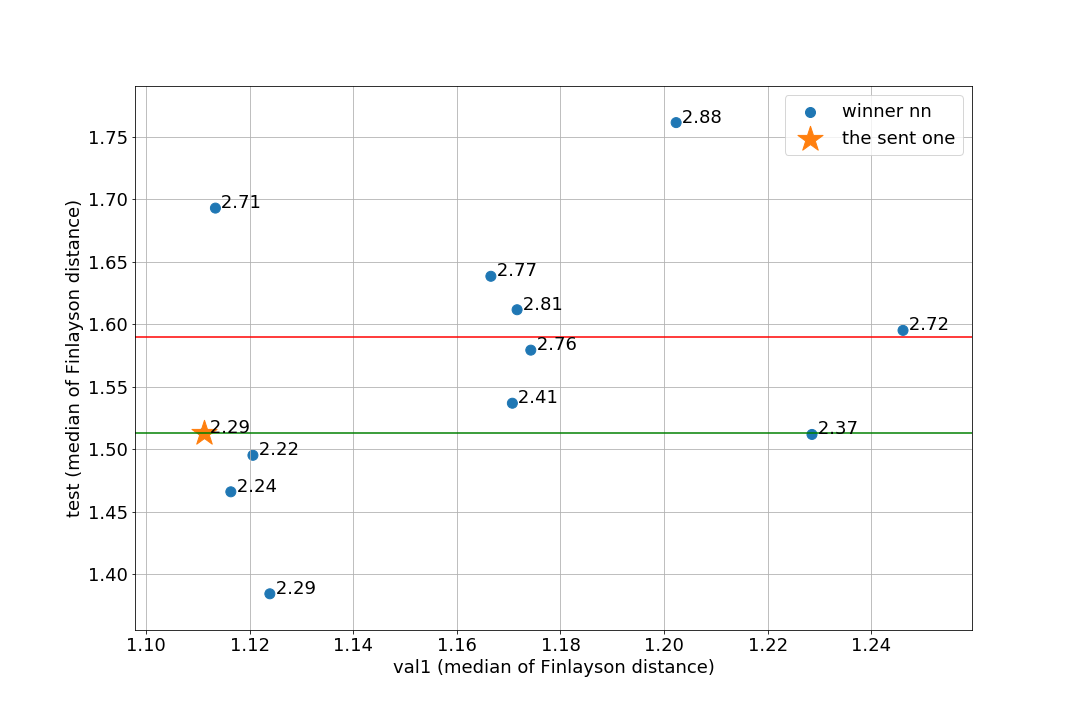}
    
    \caption{Median Finlayson distance on validation/test scatter plot for different variation of winner architecture of neural networks. 
    The red line denotes the score of the 2nd place in the contest, the green line -- the first place. 
    Numbers show average Finlayson distance.} 
    \label{fig:win_joinedval}
\end{figure}

The best result on our validation data was achieved by a convolution neural network method with a GrayWorld (see Fig. \ref{fig:win_arch}) calculated in the first layers with $L_2$-regularization parameter $r=10^{-5}$. 
As was mentioned before the network was learned to predict not only the answer (3 numbers), but also both sides of the grey cube, 9 number in total.
So, it's like 3-headed, that's why we've called this paper color cerberus.
We've trained several networks with the same parameters and choose the best one according the validation (see Fig. \ref{fig:win_joinedval}). 
Here ``Conv $3\times3\times N \rightarrow M$'' denotes convolution layer with $3 \times 3$ window size, $N$ input filter and $M$ output filter.

The best loss function happens to be the mean absolute error (MAE), that just averages all the absolute difference values:
\begin{equation*}
    MAE(T, P) = \frac{\sum _ {i, j} \left| t_i^j - p_i^j \right| }{3n},
\end{equation*}
where $T_i^j, P_i^j, i \in \{1, ..., n\}, j \in \{1,2,3\}$ are respectively ground Truth and Predicted values.
Also we've tested mean square error (MSE) and averaged Finlayson distance but validation results was worse. 

Finally, taking into account that our goal was median Finlayson error, at each learning step we've threw out outliers, precisely we've ignored the worst 20\% and the best 1\% of patches in every 256-length mini-batch.

During the contest we've tried to develop some averaging technique to use all learning results and boost quality.
We've tried three ways: simple answer averaging, weighted averaging (weights was equal to subvalidation score), searching the closest answer to the rest answers.
Unfortunately all this methods fails, so we've decided simply choose network with smallest validation value.
Indeed, it can be seen from figure (\ref{fig:win_arch}) that there is some dependency between validation and test scores.

\subsection{What does really work?}

\begin{figure}
    \centering
    \includegraphics[width=0.15\textwidth]{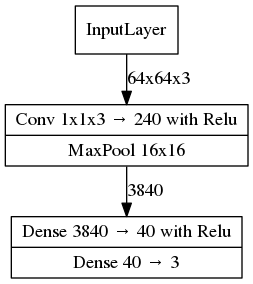}
    
    \caption{Architecture of the Bianco neural network \cite{bianco2017single}. There are just a few layers, 3 outputs and no implicit GreyWorld calculation.}
    \label{fig:italian_arch}
\end{figure}

After the competition was over, we thoroughly tested the group of methods based on the \cite{bianco2017single}. 
It was not a full-fledged reimplementation: only a local estimator was trained (see Fig. \ref{fig:italian_arch}). 

\begin{figure}[ht]
    \centering
    \includegraphics[width=0.5\textwidth]{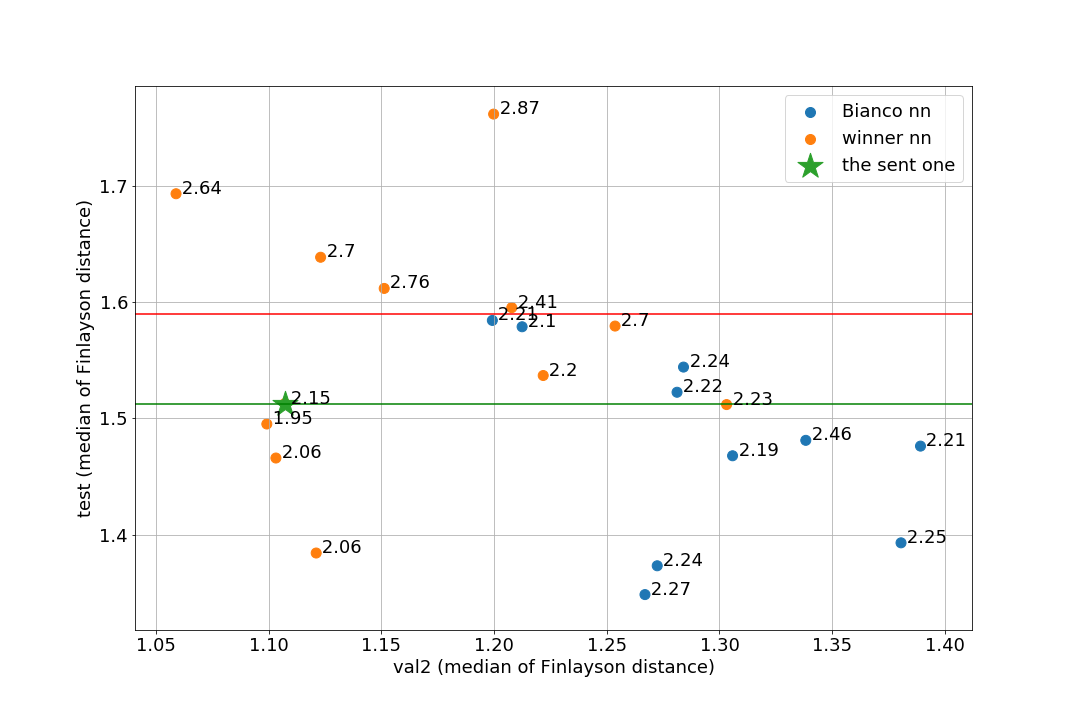}
    
    \caption{Bianco architecture \cite{bianco2017single} versus winner architecture. 
    The red line denotes the score of the 2nd place in the contest, the green line -- the first place. 
    Numbers show average Finlayson distance.} %
    \label{fig:winner_italian_val}
\end{figure}

We experimented with them during the competition, although these nets were slower to train and showed themselves worse on early validations. 
However, local estimator from \cite{bianco2017single} shows great quality with less parameter tuning \ref{fig:winner_italian_val}.

\section {Conclusion}

Despite of their widespread use, the effectiveness of multilayer CNN is still not so easy to explain. 
The most popular story -- that a well trained CNN architecture enforces some kind of hierarchical image analysis, with lower levels corresponding to local texture features and higher levels being responsible for semantics -- is not only vague and incomplete, but also partially wrong: the texture bias of CNNs is now a well established issue.

For the global illuminant estimation task this texture bias is, as it seems, not a curse but a blessing.

On the other hand, more principled analysis of local illumination histograms as performed in \cite{bianco2017single} has clear advantages. One can easily see the connections of this approach to both classical gamut estimation in color vision, and a popular PointNet architecture for 3d point cloud processing.

To make further progress the community needs better data to check applicability of more sophisticated approaches, and Cube+ effort is definitely a good step in that direction.

\section*{Affiliation}
We are very grateful to Konstantin Soshin and Arseniy Terekhin for helping with dataset collection and processing, also to Dmitry Nikolaev for fruitful advises.

The paper was partially funded by RFBR according to the research project \#17-29-03514 .

\bibliographystyle{IEEEtran}
\bibliography{IEEEabrv,biblio}

\begin{thebibliography}{10}
\providecommand{\url}[1]{#1}
\csname url@samestyle\endcsname
\providecommand{\newblock}{\relax}
\providecommand{\bibinfo}[2]{#2}
\providecommand{\BIBentrySTDinterwordspacing}{\spaceskip=0pt\relax}
\providecommand{\BIBentryALTinterwordstretchfactor}{4}
\providecommand{\BIBentryALTinterwordspacing}{\spaceskip=\fontdimen2\font plus
\BIBentryALTinterwordstretchfactor\fontdimen3\font minus
  \fontdimen4\font\relax}
\providecommand{\BIBforeignlanguage}[2]{{%
\expandafter\ifx\csname l@#1\endcsname\relax
\typeout{** WARNING: IEEEtran.bst: No hyphenation pattern has been}%
\typeout{** loaded for the language `#1'. Using the pattern for}%
\typeout{** the default language instead.}%
\else
\language=\csname l@#1\endcsname
\fi
#2}}
\providecommand{\BIBdecl}{\relax}
\BIBdecl

\bibitem{bianco2017single}
S.~Bianco, C.~Cusano, and R.~Schettini, ``Single and multiple illuminant
  estimation using convolutional neural networks,'' \emph{IEEE Transactions on
  Image Processing}, vol.~26, no.~9, pp. 4347--4362, 2017.

\bibitem{bianco2015color}
------, ``Color constancy using cnns,'' in \emph{Proceedings of the IEEE
  Conference on Computer Vision and Pattern Recognition Workshops}, 2015, pp.
  81--89.

\bibitem{colorchecker}
\BIBentryALTinterwordspacing
L.~Shi and B.~Funt. Re-processed version of the gehler color constancy dataset
  of 568 images. [Online]. Available: \url{http://www.cs.sfu.ca/~colour/data/}
\BIBentrySTDinterwordspacing

\bibitem{banic2017unsupervised}
N.~Bani{\'c} and S.~Lon{\v{c}}ari{\'c}, ``Unsupervised learning for color
  constancy,'' \emph{arXiv preprint arXiv:1712.00436}, 2017.

\bibitem{aytekin2017data}
{\c{C}}.~Aytekin, J.~Nikkanen, and M.~Gabbouj, ``A data set for
  camera-independent color constancy,'' \emph{IEEE Transactions on Image
  Processing}, vol.~27, no.~2, pp. 530--544, 2017.

\bibitem{cheng2014illuminant}
D.~Cheng, D.~K. Prasad, and M.~S. Brown, ``Illuminant estimation for color
  constancy: why spatial-domain methods work and the role of the color
  distribution,'' \emph{JOSA A}, vol.~31, no.~5, pp. 1049--1058, 2014.

\bibitem{ciurea2003large}
F.~Ciurea and B.~Funt, ``A large image database for color constancy research,''
  in \emph{Color and Imaging Conference}, vol. 2003, no.~1.\hskip 1em plus
  0.5em minus 0.4em\relax Society for Imaging Science and Technology, 2003, pp.
  160--164.

\bibitem{hemrit2018rehabilitating}
G.~Hemrit, G.~D. Finlayson, A.~Gijsenij, P.~Gehler, S.~Bianco, B.~Funt,
  M.~Drew, and L.~Shi, ``Rehabilitating the colorchecker dataset for illuminant
  estimation,'' in \emph{Color and Imaging Conference}, vol. 2018, no.~1.\hskip
  1em plus 0.5em minus 0.4em\relax Society for Imaging Science and Technology,
  2018, pp. 350--353.

\bibitem{finlayson2014reproduction}
G.~D. Finlayson and R.~Zakizadeh, ``Reproduction angular error: An improved
  performance metric for illuminant estimation,'' \emph{perception}, vol. 310,
  no.~1, pp. 1--26, 2014.

\bibitem{scikit_local_mean}
``downscale\_local\_mean,''
  \url{https://scikit-image.org/docs/dev/api/skimage.transform.html}.

\end{thebibliography}

\end{document}